\documentclass{article}

\usepackage[preprint]{corl_2026} 

\hypersetup{
  pdftitle={Selective Agentic Recovery for UAV Autonomy with a Persistent Mission Runtime},
  pdfauthor={Taewoo Park, Kyeonghyun Yoo, Seunghyun Yoo, Hwangnam Kim},
  pdfsubject={Preprint},
  pdfkeywords={robot learning, UAV autonomy, agentic AI}
}

\usepackage{amsmath}
\usepackage{amssymb}
\usepackage{booktabs}
\usepackage{graphicx}

\renewcommand{\topfraction}{0.95}
\renewcommand{\bottomfraction}{0.95}
\renewcommand{\textfraction}{0.05}
\renewcommand{\floatpagefraction}{0.85}
\title{Selective Agentic Recovery for UAV Autonomy with a Persistent Mission Runtime}

\author{
  Taewoo Park \quad Kyeonghyun Yoo \quad Seunghyun Yoo \quad Hwangnam Kim\\
  Department of Electrical and Electronic Engineering\\
  Korea University\\
  \texttt{\{taewoopark,seven1705,seunghyunyoo,hnkim\}@korea.ac.kr}\\
  {\small Corresponding author: Hwangnam Kim}
}
\date{}

\begin{document}
\maketitle

\begin{abstract}
Agentic AI can support unmanned aerial vehicle (UAV) autonomy by providing
high-level recovery reasoning when local waypoint- or setpoint-based execution
encounters blocked passages, repeated no-progress behavior, or mission-level
ambiguity. On physical UAVs, however, remote reasoning is most useful when it is
invoked selectively, since each call introduces latency, resource cost, backend
uncertainty, and a need to validate the returned decision. This paper presents
Persistent Mission Runtime (PMR), a UAV recovery framework that keeps the
mission loop and safety-critical execution local while using an external
agentic reasoner only as an on-demand recovery module. The reasoner selects
from predefined recovery skills, and each returned decision is parsed,
verified, safety-filtered, and mapped to local executor actions before it can
affect flight. PMR introduces learned Cognitive Value of Invocation
(learned-CVI), a compact admission gate that estimates when remote agentic
reasoning is likely to improve near-term mission progress enough to justify its
operational cost. Across a fixed 400-run Gazebo/PX4 benchmark with eight
scenarios, learned-CVI raises hard/ambiguous-regime success from 5.0\% under
local-only autonomy to 95.0\%, outperforms one-shot and periodic reasoning
baselines by 20.0 and 32.5 percentage points, and reduces remote-agent calls by
16.7\% and logged tokens by 29.2\% relative to a manually tuned rule-based
invocation baseline.
\end{abstract}
\section{Introduction}

In emerging agentic-AI-assisted unmanned aerial vehicle (UAV) autonomy systems,
remote agentic reasoners offer a promising way to extend local autonomy with
context-aware recovery decisions. While this paradigm is increasingly explored
as a high-level reasoning layer for robot autonomy
\citep{sapkota2025agenticuavsurvey,wang2024llmagentsurvey,
driess2023palme,zitkovich2023rt2,
koubaa2025agenticuavs},
UAVs present a particularly demanding setting; aerial missions inherently
combine long horizons, partial perception, limited onboard resources, and
recovery states that are difficult to fully enumerate in advance. Fortunately,
modern UAV flight stacks already provide robust layered multicopter control
and offboard interfaces
\citep{px4_offboard_mode}, making waypoint- or setpoint-centric autonomy
practical. These capabilities offer a strong substrate for higher-level
recovery when a UAV encounters dead ends, no-progress loops, or mission-level
ambiguity. The central challenge lies in selectively admitting remote reasoning
while keeping execution locally verified.

Selective admission is critical because each remote invocation may add latency,
token cost, remote-backend uncertainty, and validation burden. A UAV runtime
should therefore invoke remote reasoning only when the expected recovery value
justifies these costs. Once admitted, reasoner outputs must follow a
bounded execution path before impacting physical flight.
Routing foundation-model reasoner outputs as raw setpoints, velocities, or
platform-specific commands severely complicates verification and safety
filtering. To address this, the resulting runtime problem is to couple
cost-aware admission with locally verified execution, positioning remote
reasoning as a sparse recovery resource inside a selective mission layer.

Prior robot-language and UAV systems have connected language-model reasoning to
embodied action through application programming interfaces (APIs), skills,
policy programs, embodied replanning,
vision-language-action models, and reason-act-observe loops
\citep{zitkovich2023rt2,vemprala2023chatgptrobotics,
jiao2023swarmgpt,aikins2024leviosa,ahn2022saycan,huang2022inner,
shah2023lmnav,liang2023codeaspolicies,yao2023react}.
These systems make the connection between language-model reasoning and robot
action increasingly practical. However, selective invocation remains a
challenging runtime design problem. Current frameworks often invoke reasoning
at static intervals, assume an always-active reasoner, or rely on
expert-specified trigger logic. For physical UAVs, a useful admission policy should
trigger remote recovery only when the expected utility of the decision
outweighs the combined penalties of latency, token expenditure, backend
uncertainty, and downstream validation.

We present Persistent Mission Runtime (PMR), a runtime framework that turns
agentic AI into an on-demand mission agent for UAV recovery. PMR consumes
runtime observations to update mission memory and compact runtime state, then
decides whether to continue local execution or construct a compact recovery
prompt for the agentic reasoner. Prompt construction and remote-agent
invocation become runtime-managed selective admission decisions. The local
flight stack remains responsible for continuous execution, while PMR governs
sparse mission-level recovery admission.
In our implementation, OpenClaw provides the offboard agentic-AI workspace:
recovery queries are exposed as typed tools, and other mission-level tools such
as waypoint geocoding, weather/wind checks, global positioning system (GPS)
localization, navigation, and sensing/perception interfaces can share the same
locally verified PMR boundary.

PMR addresses these challenges by providing a verifiable execution path for
admitted agent decisions. The remote reasoner operates exclusively over a set
of predefined recovery skills; its outputs influence the UAV only after passing
through sequential stages of parsing, local verification, safety shielding,
fallback checks, and executor mapping. This bounded interface constrains
open-ended agent reasoning into structured recovery actions, satisfying the
rigorous safety and verifiability expectations inherent to UAV deployments.

The core decision-making problem in PMR is determining whether a remote
reasoning invocation genuinely justifies its operational costs. To address
this, we introduce the learned Cognitive Value of Invocation (learned-CVI), a
cost-aware and uncertainty-sensitive admission criterion computed over compact
runtime features. Learned-CVI evaluates whether remote agentic reasoning is
likely to improve mission progress sufficiently to outweigh the combined
overhead of latency, token expenditures, budget depletion, backend uncertainty,
and downstream validation burdens.

This paper makes three contributions.
\begin{itemize}
    \item We formulate selective remote reasoning for
    agentic-AI-assisted UAV autonomy and introduce PMR, a persistent mission
    runtime that decides between continued local autonomy and costly remote
    recovery reasoning.
    \item We introduce learned-CVI, a lightweight admission score trained from
    short-horizon recovery-utility labels, and integrate it with fixed runtime
    guards to suppress nominal over-query while preserving hard-regime
    recovery.
    \item We evaluate PMR using a fixed-protocol simulation benchmark,
    comparing it against diverse baselines including local-only, one-shot,
    periodic, rule-based, and learned invocation policies. To demonstrate
    practical viability, we complement this simulation study with a real-world
    demonstration on a Crazyflie nano-quadcopter.
\end{itemize}

The rest of the paper is organized as follows. Section~2 reviews related work.
Section~3 presents PMR and learned-CVI. Section~4 reports simulation, ablation,
and real-platform results. Section~5 discusses scope and future extensions, and
Section~6 concludes.
\section{Related Work}

PMR builds on three lines of work: agentic robot-language systems, bounded
layered autonomy, and selective reasoning under costly inference. Its focus is
the runtime admission challenge for physical UAVs: admitting a remote agentic
reasoner while maintaining bounded output authority.

\paragraph{Agentic Robot-Language Systems.}
Agentic AI systems combine reasoning, memory, tool use, perception, and
environment interaction
\citep{sapkota2025agenticuavsurvey,wang2024llmagentsurvey,
driess2023palme,koubaa2025agenticuavs}. Robot-language/UAV
systems connect these models to missions, waypoints, tools, grounded skills,
policy programs, multimodal actions, and reason-act-observe loops
\citep{zitkovich2023rt2,vemprala2023chatgptrobotics,
jiao2023swarmgpt,aikins2024leviosa,ahn2022saycan,huang2022inner,
shah2023lmnav,liang2023codeaspolicies,yao2023react,lee2026rhga,
jung2024uavswarmedgeai}. PMR treats the reasoner as an on-demand recovery
module admitted from mission state, expected recovery value, and invocation
cost.

\paragraph{Layered Autonomy with Local Verification.}
Modern UAV stacks separate high-level mission execution from low-level flight
control. PMR follows this layered design by adding a sparse recovery-admission
layer above the stabilizer, planner, and flight controller. Runtime safety
filters also separate proposed actions from execution authority
\citep{ames2017cbf,wabersich2021predictive,nguyen2024gameplay}. PMR adopts
this separation for agentic recovery: remote outputs are bounded
recovery decisions routed through local parsing, verification, shielding,
fallback handling, and execution.

\paragraph{Selective Reasoning.}
Tool-use and agent-loop methods show how models can call APIs while
interleaving reasoning with observations and actions
\citep{yao2023react,schick2023toolformer}. PMR brings this tool-use view into
a persistent UAV mission runtime: the control loop continues locally, while
learned-CVI admits bounded recovery queries only when runtime evidence
justifies invocation.

Together, these lines of work motivate a runtime paradigm for agentic UAV
autonomy where remote reasoning is a selective recovery resource admitted
through cost-sensitive decisions under local verification constraints.
\section{Method}

PMR treats agentic AI as an active recovery decision maker, not a passive
predictor. For physical UAV use, this requires three runtime roles: admission,
limited authority, and local execution. A PPO-based local execution substrate
keeps the mission moving by default; learned-CVI decides when a degraded mission
state justifies a remote recovery query; the recovery-skill contract limits what
the agent may return; and local verification maps accepted decisions to the
vehicle. Remote reasoning is most useful in blocked or ambiguous states, where
latency, budget use, and validation burden also matter most. This separation
turns agentic UAV recovery from direct control into runtime resource allocation:
PMR decides when remote reasoning is valuable and admits only a locally
checkable recovery decision.
This section formalizes the mission state, admission rule, recovery-skill
interface, and pre-deployment gate training procedure.

\subsection{Persistent Mission Runtime}

\begin{figure}[t]
    \centering
    \includegraphics[width=\linewidth]{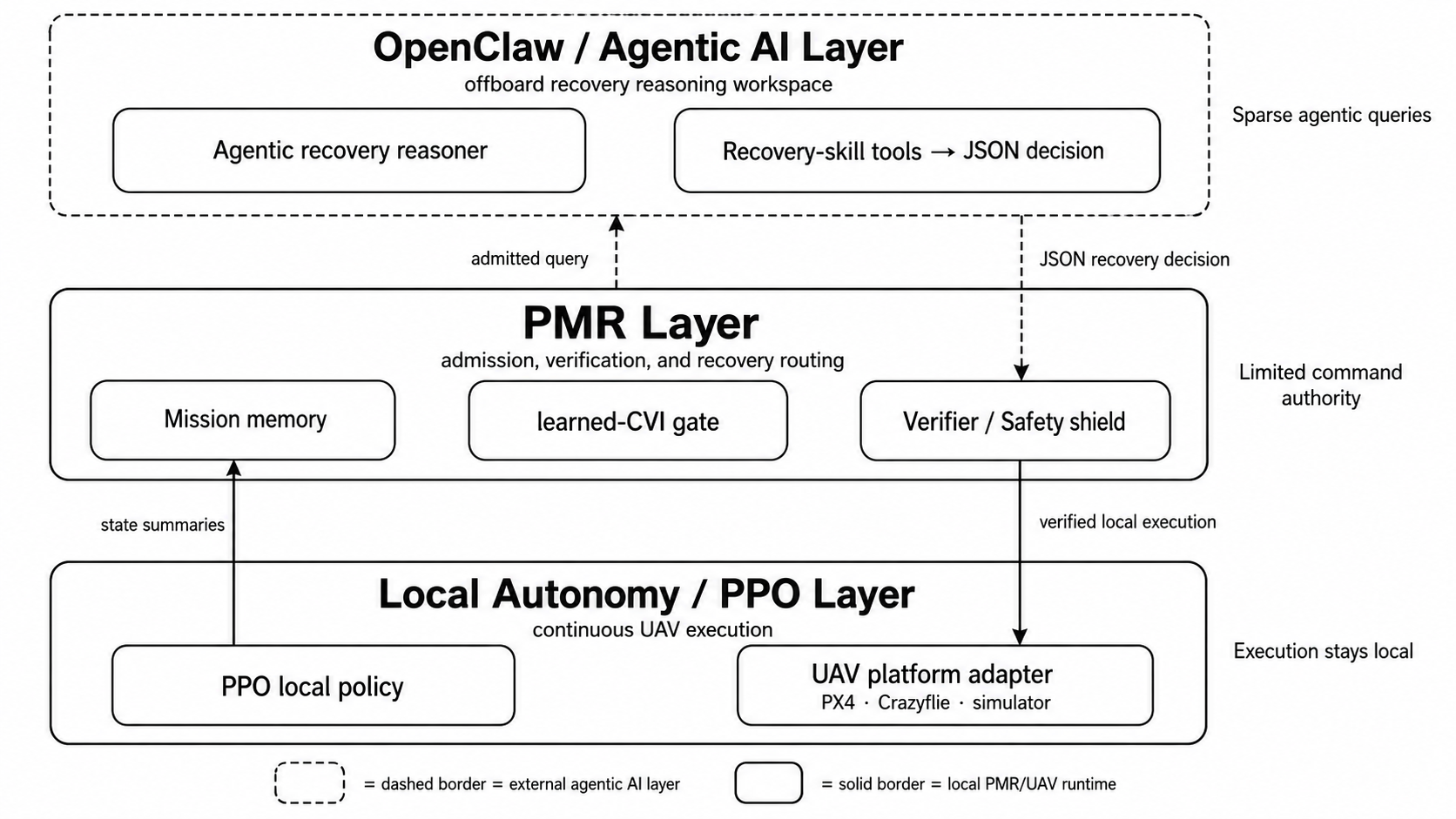}
    \caption{Layered PMR runtime. OpenClaw realizes the external agentic-AI
    workspace, while PMR owns admission, recovery-skill verification, and local
    execution.}
    \label{fig:pmr-runtime-loop}
\end{figure}

Figure~\ref{fig:pmr-runtime-loop} shows the PMR loop. PMR updates mission
memory from platform observations and uses local execution by default. When the
state indicates progress loss, ambiguity, or local recovery difficulty, PMR
constructs a compact prompt for the remote/offboard recovery reasoner.
In our implementation, OpenClaw provides this offboard agentic-AI workspace and
tool-calling boundary, while PMR keeps the mission loop, verification, and
execution local. Recovery selection is the tool used in this paper, but the
same boundary can also host typed mission tools such as geocoding a natural
language destination into a waypoint, checking weather context, reading GPS or
tracking state, and summarizing sensing/perception cues.

We write the PMR runtime state and update rule as
\begin{equation}
    x_t=(m_t,v_t,b_t,\eta_t),\qquad
    x_{t+1}=F_{\mathrm{PMR}}(x_t,o_t,r_t),
\end{equation}
where \(m_t\) is mission memory, \(v_t\) is verifier state, \(b_t\) is the
remaining query budget, \(\eta_t\) collects executor/logging state, \(o_t\) is
the platform observation, and \(r_t\) is the local or verified recovery
decision applied at the update.
This persistent state makes each remote call an online mission decision:
observations update evidence for local continuation or recovery admission.

\subsection{Selective Reasoning Admission}

PMR treats remote reasoning as an online admission problem above a competent
local stack. Runtime observations update compact mission state, and admitted
recovery intent becomes locally verified executor action.

Let \(\mathcal{O}\) denote the platform observation space, \(\mathcal{S}\) the
compact PMR runtime-state space, \(\mathcal{A}_{\mathrm{loc}}\) the local
executor action space, and \(\mathcal{D}_{\mathrm{rec}}\) the predefined
recovery-skill space. PMR receives \(o_t\in\mathcal{O}\), constructs
\(s_t=\phi(x_t,o_t)\in\mathcal{S}\), and applies an admission policy
\(\pi_{\mathrm{adm}}:\mathcal{S}\rightarrow\{0,1\}\). This binary decision
selects local continuation or verified recovery:
\begin{equation}
    \pi_{\mathrm{adm}}:s_t\mapsto q_t\in\{0,1\},\qquad
    u_t=
    \begin{cases}
        \pi_{\mathrm{loc}}(s_t), & q_t=0,\\
        \mathcal{E}(\rho_t), & q_t=1,
    \end{cases}
\end{equation}
where \(\pi_{\mathrm{loc}}\) is the fixed local policy, \(\rho_t\) is a
verified recovery decision, and \(\mathcal{E}\) maps that decision to the local
executor.

Our implementation uses a Proximal Policy Optimization (PPO)-trained
flying-phase policy \citep{schulman2017ppo} with a 10-dimensional normalized
observation and \(\mathrm{Discrete}(32)\) tool-repeat macro actions. PPO gives
a repeatable local policy; PMR can use other local stacks. All evaluation
policies share this local policy, verifier, shield, and executor; only the
admission rule changes. This keeps the comparison focused on which runtime
states justify a remote recovery query.

\subsection{Recovery-Skill Execution Boundary}

Raw agent outputs create a verification problem because their command space can
be open-ended. PMR constrains the reasoner to a closed recovery-skill
vocabulary, turning agent output into a typed, locally checkable decision
instead of a direct flight command. The PMR contract limits returned JSON to
recovery-skill decisions.

Let \(\mathcal{D}_{\mathrm{rec}}\) denote the predefined recovery-skill set and
let \(y_t\) be the returned JSON object. PMR admits an agent output only through
\begin{equation}
    \rho_t =
    \mathcal{S}_{\mathrm{safe}}\!\left(
    \mathcal{V}_{\mathrm{loc}}\!\left(
    \mathcal{P}_{\mathrm{json}}(y_t)\right)\right)
    \in \mathcal{D}_{\mathrm{rec}}\cup\{\bot\},
\end{equation}
where parsing, local verification, and shielding produce either a predefined
recovery skill or a local fallback \(\bot\).

The schema encodes recovery skills and a flight-stack command boundary;
Appendix~\ref{app:reasoner-contract} reports the checked schema. PMR fills the
prompt with compact mission state, perception summaries, budget/query state,
and recovery options. Returned JSON is parsed, verified, shielded, and mapped
to the executor; parser, safety, timeout, and infrastructure events use local
fallback behavior. The remote agent contributes recovery choice, while PMR
keeps local control over timing, verification, and platform execution.

\subsection{CVI Admission Score}

Telemetry thresholds alone provide an incomplete signal for when a reasoning
call is valuable: a blocked heading matters most after local repair has stopped
making progress. Instead of predicting task success directly, learned-CVI
estimates whether remote reasoning should improve short-horizon recovery
utility relative to continuing locally.
The useful signal is not obstacle proximity in isolation, but proximity
conditioned on stalled local progress, remaining query budget, recent recovery
attempts, and verifier/executor state.
Learned-CVI implements the utility comparison in Section~3.5 as an online
admission gate trained before deployment and held fixed during deployment.
The gate computes
\begin{equation}
    z_t = w^\top \mathrm{norm}(\phi(s_t)) + b,\qquad
    c_t = \mathrm{CVI}(s_t) = \sigma(z_t),
\end{equation}
where \(\phi(s_t)\) is a fixed 18-dimensional (18D) runtime feature vector from
telemetry, executor feedback, verifier state, local-planner state, budget/query
state, and compact perception summaries. The groups have clear runtime meaning:
progress asks whether local execution is working; risk/ambiguity asks whether
local repair is reliable; budget/query asks whether a call is affordable; and
verifier/executor state records recent recovery outcomes. PMR learns from these
semantic summaries instead of raw sensor arrays, making the gate less tied to a
specific sensing payload and easier to reuse across simulator range cues and
small-UAV bottom range/optical-flow cues.
Appendix~\ref{app:cvi-details} gives the full feature contract.

PMR combines this learned score with fixed runtime guards before issuing a
reasoner request. Let \(g_t\in\{0,1\}\) denote fixed runtime guards,
including available query budget, cooldown, and terminal-radius suppression,
and let \(h_t\in\{0,1\}\) denote the pre-specified hard-stuck guard. The deployed query
decision is
\begin{equation}
    q_t =
    \mathbb{I}\!\left[
        g_t = 1 \;\wedge\; \left(c_t \ge \tau_{\mathrm{lite}} \;\vee\; h_t = 1\right)
    \right].
\end{equation}
In the final evaluation, learned-CVI uses \(\tau_{\mathrm{lite}}=0.997\), fixed
budget/cooldown/terminal guards, and a pre-specified hard-stuck guard triggered
only by no-progress and blocked-motion. The threshold reflects the calibrated
score scale of this conservative linear gate.

\subsection{Pre-Deployment Gate Training}

The learned-CVI gate is trained before deployment from PMR logs and held fixed during
deployment. For each logged state, PMR uses a \(K=5\) short-horizon label:
it compares outcomes over the next five runtime steps, and invocation is useful
when it yields more progress than local continuation after token, latency,
safety-shield, and planner-difficulty costs.
Formally, the short-horizon recovery utility is
\begin{equation}
\begin{aligned}
    U_t^{\mathrm{inv}} &=
    \Delta p_t^{\mathrm{inv}}
    - \lambda_{\mathrm{tok}} C_t^{\mathrm{tok}}
    - \lambda_{\mathrm{lat}} C_t^{\mathrm{lat}} \\
    &\quad
    - \lambda_{\mathrm{safe}}\mathbb{I}[\mathrm{shield}_t]
    - \lambda_{\mathrm{fail}}\mathbb{I}[\mathrm{fail}_t], \\
    \ell_t &=
    \mathbb{I}\!\left[
    U_t^{\mathrm{inv}} - U_t^{\mathrm{loc}} > 0
    \right].
\end{aligned}
\end{equation}
Here \(U_t^{\mathrm{loc}}\) is local-continuation utility,
\(\Delta p_t^{\mathrm{inv}}\) is short-horizon progress after invocation, and
\(\ell_t\) is the recovery-utility label. The CVI gate estimates the probability that
invocation utility exceeds local continuation. The short-horizon label is
intentional: the gate estimates near-term recovery value, not long-horizon
mission success, which would mix invocation quality with local-policy quality.
The gate is re-evaluated throughout the mission loop. The training split
contains 651 labeled states after
filtering; 404 ambiguous rows are kept outside binary supervision. A separate
533-row validation split selects the threshold before the fixed-protocol
400-run simulation evaluation.

Because the deployed query rule combines a learned score with fixed guards, the
results report guard-only, learned-CVI score-only, and matched-budget
rule-based ablations. Additional appendix analyses characterize the design
without changing the deployed gate or evaluation protocol. Thus learned-CVI supplies the
admission score, while the runtime contract keeps recovery execution
structured, locally checkable, and comparable.
\section{Experimental Results}
\label{sec:result}

\subsection{Evaluation Protocol}

We evaluate PMR with a fixed-protocol simulation benchmark that isolates the
mission-level query rule while holding the runtime contract fixed. Here, large
language model (LLM) denotes the shared remote reasoner. The simulation uses a
software-in-the-loop flight-stack backend, MAVSDK execution, compact
telemetry/range summaries, and an OpenClaw HTTP JSON gateway for
limited-authority agent calls. We evaluate eight scenarios with ten seeds each,
grouped before reporting into nominal and hard/ambiguous regimes. The policies
are local-only, one-shot LLM, periodic LLM, rule-based LLM, and learned-CVI. All
methods share the runtime, reasoner contract, verifier, safety shield, timeout,
and logging schema; only the admission policy changes. Clean Success@1m
requires final distance at most 1m with no collision, z violation, or backend
failure.
Appendices~\ref{app:repro}--\ref{app:rule-based} provide reproducibility,
protocol, scenario, local-policy, and rule-trigger details.

\subsection{Benchmark Performance}

\begin{table}[t]
\centering
\small
\setlength{\tabcolsep}{2.4pt}
\renewcommand{\arraystretch}{0.92}
\caption{Fixed-protocol 400-run simulation evaluation.}
\label{tab:fixed-protocol-eval}
\vspace{-5pt}
\begin{tabular*}{\linewidth}{@{\extracolsep{\fill}}llrrrrrr@{}}
\toprule
Regime & Method & Runs & Clean Success@1m & Rate & Calls & Tokens & Backend fail \\
\midrule
Nominal & local-only & 40 & 35 & 0.875 & 0.000 & 0.0 & 0 \\
 & one-shot LLM & 40 & 39 & 0.975 & 1.000 & 571.6 & 1 \\
 & periodic LLM & 40 & 34 & 0.850 & 3.000 & 1675.8 & 6 \\
 & rule-based LLM & 40 & 37 & 0.925 & 1.200 & 687.5 & 3 \\
\addlinespace[1pt]
 & \textbf{learned-CVI (ours)} & 40 & \textbf{40} & \textbf{1.000} & \textbf{0.150} & \textbf{60.3} & \textbf{0} \\
\midrule
Hard/ambiguous & local-only & 40 & 2 & 0.050 & 0.000 & 0.0 & 0 \\
 & one-shot LLM & 40 & 30 & 0.750 & 1.000 & 443.4 & 2 \\
 & periodic LLM & 40 & 25 & 0.625 & 3.000 & 1194.8 & 9 \\
 & rule-based LLM & 40 & \textbf{39} & \textbf{0.975} & 1.200 & 683.0 & \textbf{1} \\
\addlinespace[1pt]
 & \textbf{learned-CVI (ours)} & 40 & 38 & 0.950 & \textbf{1.000} & \textbf{483.4} & \textbf{1} \\
\bottomrule
\end{tabular*}
\vspace{-6pt}
\end{table}

Table~\ref{tab:fixed-protocol-eval} reports the 400-run simulation evaluation.
In the hard/ambiguous regime, local-only reaches 2/40 Clean Success@1m,
one-shot LLM 30/40, periodic LLM 25/40, rule-based LLM 39/40, and learned-CVI
38/40. The rule-based baseline represents expert recovery-trigger knowledge.
Learned-CVI nearly matches it with a learned value signal, showing that much of
the trigger logic can be recovered from compact runtime evidence.

\begin{table}[t]
\centering
\small
\setlength{\tabcolsep}{2.4pt}
\renewcommand{\arraystretch}{0.92}
\caption{Per-scenario learned-CVI results.}
\label{tab:per-world-learned-cvi}
\vspace{-5pt}
\begin{tabular*}{\linewidth}{@{\extracolsep{\fill}}llrrrrr@{}}
\toprule
Scenario & Regime & Runs & Clean Success@1m & Final dist. (m) & Calls & Tokens \\
\midrule
Empty & Nominal & 10 & 10 & 0.585 & 0.0 & 0.0 \\
Static Obstacles &  & 10 & 10 & 0.609 & 0.0 & 0.0 \\
Mild Random Obstacles &  & 10 & 10 & 0.607 & 0.6 & 241.1 \\
Recoverable Dead End &  & 10 & 10 & 0.539 & 0.0 & 0.0 \\
\midrule
Multi-Obstacle & Hard/ambiguous & 10 & 10 & 0.629 & 1.0 & 484.9 \\
Hard Re-entry A &  & 10 & 10 & 0.601 & 1.0 & 601.8 \\
Hard Re-entry B &  & 10 & 9 & 1.832 & 0.9 & 243.0 \\
Dead-end Bypass &  & 10 & 9 & 0.731 & 1.1 & 604.1 \\
\bottomrule
\end{tabular*}
\vspace{-6pt}
\end{table}

Table~\ref{tab:per-world-learned-cvi} gives the per-scenario learned-CVI
breakdown. The two remaining hard/ambiguous cases occur in Hard Re-entry B and
Dead-end Bypass, matching the under-query and backend-timeout diagnostics in
Section~\ref{sec:cvi-ablation}.

\subsection{Invocation Efficiency}

\begin{figure}[t!]
    \centering
    \includegraphics[width=\linewidth]{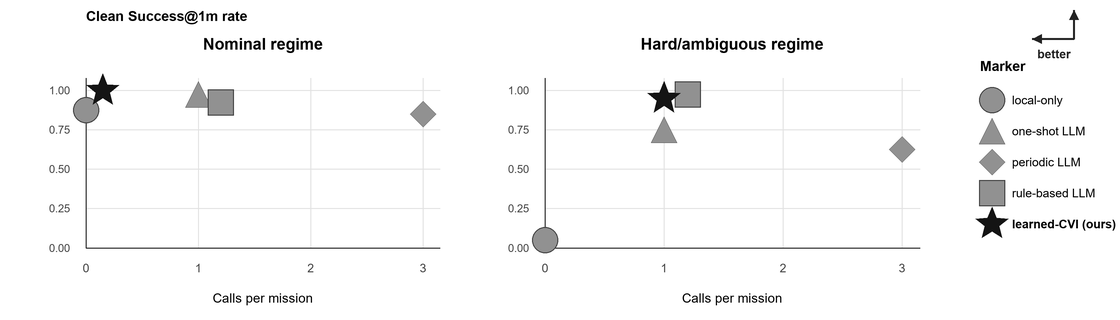}
    \caption{Success/call trade-off by regime: nominal (left) and
    hard/ambiguous (right).}
    \label{fig:success-resource}
    \vspace{-8pt}
\end{figure}

Figure~\ref{fig:success-resource} visualizes both regimes. In nominal settings,
learned-CVI reaches 40/40 Clean Success@1m with 0.150 calls, using 87.5\% fewer
calls and 91.2\% fewer tokens than rule-based. In hard/ambiguous settings, it
uses 16.7\% fewer calls and 29.2\% fewer tokens while staying within one Clean
Success@1m run. PMR suppresses calls when local execution is enough and
concentrates queries when recovery is needed. Appendix~\ref{app:latency-backend}
reports latency/backend diagnostics.

\subsection{CVI Ablation Diagnostics}
\label{sec:cvi-ablation}

Table~\ref{tab:sparse-invocation-diagnostics} isolates the learned score from
runtime guards. Guard-only and learned-CVI score-only variants remain below the
full gate, while their combination gives the strongest learned sparse-query
variant; Appendix~\ref{app:sparse-invocation-diagnostics} gives counting notes.

\begin{table}[t]
\centering
\small
\setlength{\tabcolsep}{4pt}
\renewcommand{\arraystretch}{0.92}
\caption{Hard/ambiguous-regime sparse-invocation ablations.}
\label{tab:sparse-invocation-diagnostics}
\vspace{-5pt}
\begin{tabular*}{\linewidth}{@{\extracolsep{\fill}}lrrrr}
\toprule
Ablation policy & Runs & Clean Success@1m & Rate & Calls \\
\midrule
hard-stuck guard only & 40 & 16 & 0.400 & 0.400 \\
learned-CVI score only, no guard & 40 & 25 & 0.625 & 0.975 \\
matched-budget rule-based LLM & 40 & 37 & 0.925 & \(\approx 1.0\) \\
learned-CVI & 40 & 38 & 0.950 & 1.000 \\
\bottomrule
\end{tabular*}
\vspace{-6pt}
\end{table}

The deployed behavior comes from the learned score interacting with the runtime
guard structure.

\begin{table}[t]
\centering
\small
\setlength{\tabcolsep}{4pt}
\renewcommand{\arraystretch}{0.92}
\caption{CVI model-class diagnostic.}
\label{tab:cvi-model-class-main}
\vspace{-5pt}
\begin{tabular*}{\linewidth}{@{\extracolsep{\fill}}llcc}
\toprule
Split & Model & ROC-AUC & PR-AUC \\
\midrule
Validation & learned-CVI & 0.9996 & 0.9996 \\
Validation & small MLP [32,16] & 1.0000 & 1.0000 \\
Validation & boosted decision stumps & 1.0000 & 1.0000 \\
\midrule
Held-out diagnostic & learned-CVI & 0.9635 & 0.9986 \\
Held-out diagnostic & small MLP [32,16] & 0.6968 & 0.9861 \\
Held-out diagnostic & boosted decision stumps & 0.2976 & 0.9610 \\
\bottomrule
\end{tabular*}
\vspace{-6pt}
\end{table}

Table~\ref{tab:cvi-model-class-main} compares learned-CVI with a multilayer
perceptron (MLP) and gradient-boosted stumps using the same 18D features.
Receiver operating characteristic area under the curve (ROC-AUC) and
precision-recall area under the curve (PR-AUC) are threshold-free ranking
metrics. The comparison tests whether onboard-feasible lightweight models can
capture the admission signal. learned-CVI gives the best deployment trade-off:
stronger held-out ranking, transparent weights, forward-only evaluation, and
stable integration with fixed guards. Gradient-sensitivity triggers are possible
alternatives, but add online sensitivity computation and can amplify feature
noise. Appendices~\ref{app:cvi-model-diagnostic} and
\ref{app:utility-sensitivity} report data-size and utility-weight diagnostics.

\begin{table}[t]
\centering
\small
\setlength{\tabcolsep}{3.4pt}
\renewcommand{\arraystretch}{0.92}
\caption{learned-CVI threshold exposure; rates are score-crossing episodes, not
Clean Success@1m.}
\label{tab:cvi-threshold-diagnostic}
\vspace{-5pt}
\begin{tabular*}{\linewidth}{@{\extracolsep{\fill}}rcccc}
\toprule
Threshold & Nominal query eps. & Hard query eps. & Hard query exposure & Mean trigger steps/eps. \\
\midrule
0.990 & 0/30 & 23/30 & 0.767 & 6.500 \\
0.995 & 0/30 & 23/30 & 0.767 & 5.233 \\
0.997 & 0/30 & 18/30 & 0.600 & 0.600 \\
0.998 & 0/30 & 0/30 & 0.000 & 0.000 \\
\bottomrule
\end{tabular*}
\vspace{-6pt}
\end{table}

Table~\ref{tab:cvi-threshold-diagnostic} reports query-exposure rates, not
Clean Success@1m. At \(\tau_{\mathrm{lite}}=0.997\), nominal exposure is 0/30
and hard/ambiguous exposure is 18/30, showing regime-selective invocation; the
0.998 row marks the conservative operating edge. The two remaining
learned-CVI hard/ambiguous non-successes are one near-threshold under-query and
one remote-backend timeout after valid query admission. Thus, they do not
indicate an unsafe recovery decision.

\subsection{Real-Platform Demonstration}

We demonstrate PMR on a Crazyflie nano-quadcopter in blocked navigation with
the same OpenClaw-backed reasoner gateway. Table~\ref{tab:crazyflie-feasibility}
reports ten trials per policy; Appendix~\ref{app:real-platform} describes the setup.

\begin{table}[t!]
\centering
\small
\setlength{\tabcolsep}{3pt}
\renewcommand{\arraystretch}{0.92}
\caption{Crazyflie blocked-navigation real-platform result.}
\label{tab:crazyflie-feasibility}
\vspace{-5pt}
\begin{tabular*}{\linewidth}{@{\extracolsep{\fill}}lrrrrr@{}}
\toprule
Method & \begin{tabular}{@{}c@{}}Clean\\Success@1m\end{tabular} & \begin{tabular}{@{}c@{}}Final\\dist. (m)\end{tabular} & Calls & Tokens & \begin{tabular}{@{}c@{}}Safety\\events\end{tabular} \\
\midrule
local-only & 0/10 & 1.1577 & 0.0 & 0.0 & 0 \\
learned-CVI & 10/10 & 0.1922 & 1.0 & 378.1 & 0 \\
\bottomrule
\end{tabular*}
\vspace{-6pt}
\end{table}

The local-only policy obtains 0/10 Clean Success@1m at 1.1577m mean final
distance. PMR with learned-CVI-guided admission reaches 10/10 Clean Success@1m
with one recovery-skill query per trial, 0.1922m final distance, 378.1 tokens,
and zero safety events. Crazyflie's bottom range and optical-flow cues map to
semantic summaries rather than raw arrays, supporting PMR as an agent-for-UAV
runtime across sensing payloads.
\section{Limitations}

PMR is validated for single-UAV navigation recovery using semantic summaries:
progress, blocked motion, risk, query state, and verifier feedback. This tests
remote recovery admission while execution stays local. Future work should
validate the same admission, authority, and execution separation under outdoor
perception, wind, heterogeneous airframes/sensors, richer skills, and multi-UAV
operation. The two non-success cases motivate adaptive thresholds, fallback
escalation, communication-aware scheduling, and backend redundancy. Crazyflie
shows transfer through semantic summaries; larger UAVs and multi-UAV studies are
next.
\section{Conclusion}

PMR treats remote reasoning as a limited-authority recovery module for
agentic-AI-assisted UAV autonomy. Its learned-CVI gate triggers costly calls, and
accepted outputs stay locally verified. In 400-run simulation, PMR raises
hard/ambiguous Clean Success@1m from 5.0\% to 95.0\%, outperforms
one-shot/periodic reasoning, nearly matches rule-based with fewer calls/tokens,
and shows Crazyflie feasibility. With OpenClaw mission tools, PMR becomes a
practical agent-for-UAV framework for safe, efficient, extensible autonomy.

\bibliography{refs}

\clearpage
\appendix
\section{Appendix}
\setlength{\textfloatsep}{8pt plus 2pt minus 2pt}
\setlength{\floatsep}{6pt plus 2pt minus 2pt}
\setlength{\intextsep}{6pt plus 2pt minus 2pt}
\renewcommand{\topfraction}{0.95}
\renewcommand{\bottomfraction}{0.95}
\renewcommand{\textfraction}{0.05}
\renewcommand{\floatpagefraction}{0.85}
\setcounter{topnumber}{10}
\setcounter{bottomnumber}{2}
\setcounter{totalnumber}{12}

\subsection{OpenClaw Agentic-AI Contract}
\label{app:reasoner-contract}

All LLM-assisted policies use the same compact reasoner contract, parser,
allowed recovery options, timeout handling, token logging, and fallback
behavior. PMR implements the agentic-AI side with OpenClaw: the UAV runtime
exposes a recovery workspace tool that receives compact mission state, budget
status, verifier state, and the allowed recovery-skill set. In the evaluated
deployment, the OpenClaw--PMR boundary is realized through an HTTP JSON gateway
backed by the configured remote reasoner. The OpenClaw tool acts as an offboard
recovery-mode selector. It may choose one recovery skill and return typed JSON,
while PMR performs parsing, verification, safety shielding, fallback handling,
and executor mapping before the decision affects the UAV. The appendix reports
sanitized interface fields; infrastructure-specific credentials, endpoint
addresses, host identifiers, user names, and local paths are kept outside the
paper artifact.

The same OpenClaw workspace abstraction can host additional mission-level UAV
tools. For example, a natural-language destination tool can use geocoding and
map APIs to convert a place name into GPS waypoint coordinates, while a weather
tool can query rain or wind near the current UAV position. PMR treats these
outputs like recovery decisions: typed mission-level proposals that must pass
through local verification and executor mapping before they affect the UAV.
Table~\ref{tab:openclaw-pmr-tool} summarizes the recovery tool boundary and the
tool-family extension points used to interpret OpenClaw as an agent-for-UAV
workspace.

\vfill
\begin{center}
\small
\refstepcounter{table}\label{tab:openclaw-pmr-tool}
\textbf{Table~\thetable:} OpenClaw-facing PMR recovery tool interface.
\vspace{4pt}

\begin{tabular}{p{0.25\linewidth}p{0.67\linewidth}}
\toprule
Item & Sanitized implementation detail \\
\midrule
Tool role & OpenClaw exposes PMR as a recovery-query tool for offboard
agentic reasoning. The tool is called only after learned-CVI admits a query. \\
Tool input & Compact mission state, progress/no-progress evidence,
budget/cooldown status, verifier/executor state, compact sensing summaries,
and the allowed recovery-skill list. \\
Agent affordance & The agent may select one predefined recovery skill and
provide a short reason, risk estimate, and confidence value. \\
Tool output & Typed JSON returned to PMR. The output is parsed as recovery
intent rather than a direct flight command. \\
Local acceptance & PMR accepts the tool output only after parser, verifier,
safety-shield, fallback, and executor checks. \\
Extensible workspace & The same OpenClaw boundary can register weather, GPS,
localization, tracking, navigation, sensing, perception, and verified
executor-facing control tools as typed mission-level interfaces. \\
\bottomrule
\end{tabular}
\end{center}

\clearpage
\begin{table}[!t]
\centering
\small
\caption{OpenClaw workspace tool families supported by the PMR boundary.}
\label{tab:openclaw-tool-families}
\begin{tabular}{p{0.32\linewidth}p{0.58\linewidth}}
\toprule
Tool family & Example OpenClaw tools \\
\midrule
Weather context & \texttt{weather\_monitor\_tool} \\
GPS/localization & \texttt{gps\_state\_tool}, \texttt{gps\_localization\_tool} \\
Tracking/navigation & \texttt{position\_tracking\_tool}, \texttt{navigation\_tool} \\
Sensing & \begin{tabular}[t]{@{}l@{}}
\texttt{lidar\_sensing\_tool}, \texttt{marker\_sensing\_tool} \\
\texttt{crazyflie\_flowdeck\_tool}
\end{tabular} \\
Perception ingest & \texttt{sensor\_ingest\_tool}, \texttt{perception\_tool} \\
Executor-facing control & \texttt{flight\_control\_tool} \\
\bottomrule
\end{tabular}
\end{table}

\begin{table}[!t]
\centering
\small
\caption{Sanitized high-level reasoner contract for recovery-skill decisions.}
\label{tab:reasoner-contract}
\begin{tabular}{p{0.25\linewidth}p{0.67\linewidth}}
\toprule
Field & Contract content \\
\midrule
Required JSON fields & \texttt{decision}, \texttt{reason},
\texttt{suggested\_option}, \texttt{risk}, \texttt{confidence} \\
Allowed decisions &
predefined recovery-skill labels for local continuation, goal resume, goal
alignment, local repair, safe hold/verify, terminal homing, fallback-safe, and
abort-if-unsafe behavior. \\
Command boundary & Accepted outputs are structured recovery-skill choices.
Direct flight-stack commands, system commands, credentials, endpoint
addresses, and user paths remain outside the reasoner contract. \\
Failure handling & parser rejection, safety-shield rejection, verifier
rejection, timeout fallback, continued local execution, or safe hold. \\
Logged metadata & parsed decision, logged token cost, latency,
infrastructure error, post-call progress. \\
\bottomrule
\end{tabular}
\end{table}

\subsection{Learned-CVI Feature, Split, and Diagnostic Details}
\label{app:cvi-details}

Table~\ref{tab:cvi-features} reports the fixed learned-CVI runtime feature
vector. The 18-dimensional (18D) vector is derived from platform-adapter
telemetry, executor feedback, verifier state, budget/query state, and compact
perception summaries; it is an internal normalized feature contract. CVI inputs
use semantic runtime summaries and do not include scenario IDs, map templates,
or raw range/perception arrays.
Table~\ref{tab:cvi-data-split} reports the train/validation/final separation
used for the deployed learned-CVI gate.

\begin{table}[!t]
\centering
\small
\caption{learned-CVI uses a fixed normalized runtime feature vector.}
\label{tab:cvi-features}
\begin{tabular}{p{0.22\linewidth}p{0.70\linewidth}}
\toprule
Feature block & Runtime features \\
\midrule
Kinematics & normalized position \(x,y,z\) and speed. \\
Progress & normalized goal distance, waypoint progress, progress rate, and no-progress time. \\
Risk/planner & obstacle risk, target confidence, local-planner failure, and uncertainty score. \\
Query/budget & time since last query, remaining budget, and reasoning debt. \\
Runtime state & battery, previous command success, and normalized safety-intervention count. \\
\bottomrule
\end{tabular}
\end{table}

\begin{table}[!t]
\centering
\small
\caption{learned-CVI learning and evaluation separation. Training and validation
logs are used for the gate and threshold selection; the fixed-protocol 400-run
simulation aggregate is used only for final method comparison.}
\label{tab:cvi-data-split}
\begin{tabular}{p{0.16\linewidth}p{0.30\linewidth}p{0.42\linewidth}}
\toprule
Role & Source & Use and content \\
\midrule
Train &
compact-runtime train log &
fit weights and normalization from 651 labels: 283 positive and 368 negative;
404 ambiguous rows are skipped. \\
Validation &
\texttt{val} compact-runtime log &
select deployment threshold from 533 rows across 16 episodes. \\
Evaluation &
8 scenarios $\times$ 5 policies $\times$ 10 seeds &
final method comparison over 400 simulation runs. \\
\bottomrule
\end{tabular}
\end{table}

\clearpage
\begin{table}[!t]
\centering
\small
\caption{Evidence for learned-CVI as a data-driven admission gate.}
\label{tab:cvi-data-driven-evidence}
\begin{tabular}{p{0.22\linewidth}p{0.70\linewidth}}
\toprule
Evidence axis & What is fixed or learned \\
\midrule
Supervision & 651 filtered runtime states: 283 positive and 368 negative
invocation labels; 404 ambiguous rows are reserved outside binary supervision. \\
Train/validation separation & Model weights and normalization are fit on the
training split; the deployment threshold is selected on a separate validation
split with 533 rows from 16 episodes. \\
Final-evaluation separation & The 400-run simulation aggregate is reserved for final comparison after
the model, threshold, features, prompt contract, shield, verifier, scenarios,
seeds, timeout, and logging schema are fixed. \\
Input contract & The score uses compact runtime features while excluding
scenario identifiers, map templates, raw range/perception arrays, prompt text,
and OpenClaw prompt geometry. \\
Learned score behavior & Pre-deployment sensitivity analyses identify confidence,
distance/progress, no-progress, and budget/state signals as influential;
Table~\ref{tab:cvi-model-class-diagnostic} compares the deployed linear gate
with nonlinear alternatives using the same labels. \\
Closed-loop separation & The main results report guard-only, score-only, and
matched-budget rule-based LLM ablations to show how the learned score works with
fixed runtime guards. \\
\bottomrule
\end{tabular}
\end{table}

Table~\ref{tab:cvi-feature-sensitivity} summarizes the corresponding
feature-level sensitivity diagnostic. The strongest aggregate perturbation
effects come from progress/no-progress and perception-confidence/risk features,
supporting the interpretation that learned-CVI responds to reusable runtime
evidence instead of scenario labels.

\begin{table}[!t]
\centering
\small
\caption{learned-CVI feature-group sensitivity diagnostic over logged states.}
\label{tab:cvi-feature-sensitivity}
\begin{tabular*}{\linewidth}{@{\extracolsep{\fill}}lrrrr}
\toprule
Feature group & Features & Sum \(|\Delta\mathrm{CVI}|\) & Avg. \(|\Delta\mathrm{CVI}|\) & Max flip rate \\
\midrule
Progress/no-progress & 4 & 0.227 & 0.057 & 0.009 \\
Perception confidence/risk & 4 & 0.164 & 0.041 & 0.009 \\
Position/kinematics & 4 & 0.116 & 0.029 & 0.016 \\
Runtime/safety state & 3 & 0.053 & 0.018 & 0.024 \\
Query/budget & 3 & 0.045 & 0.015 & 0.009 \\
\bottomrule
\end{tabular*}
\end{table}

\subsection{Reproducibility Notes}
\label{app:repro}

The main runtime comparisons use the evaluation runner with the fixed local
policy and safety verifier held fixed. The reported aggregate contains all
requested scenario/method/seed cells for the five evaluated policies, including
one-shot LLM planning.

Reasoner calls are enabled only after preflight validation. Tokens are reported
as the logged PMR token-cost field used consistently across methods. Sanitized
contracts and aggregate token accounting are reported below; infrastructure
credentials and raw endpoint payloads remain outside the paper artifact.

\subsection{Fixed Protocol Parameters}
\label{app:parameter-freeze}

Table~\ref{tab:parameter-freeze} summarizes the fixed protocol choices used to
interpret the final 400-run evaluation. These settings define the evaluated PMR
benchmark and keep method comparisons controlled.

\begin{table}[!t]
\centering
\small
\caption{Fixed protocol parameters for the simulation comparison.}
\label{tab:parameter-freeze}
\begin{tabular}{p{0.28\linewidth}p{0.64\linewidth}}
\toprule
Parameter & Fixed setting \\
\midrule
Mission timeout & 360s shared mission budget. \\
Clean Success@1m & final distance \(\le\)1m with no collision, z violation, or backend failure. \\
Seeds and scenarios & seeds 0--9 over eight simulation scenarios. \\
Regime grouping & 4 Nominal and 4 Hard/ambiguous scenarios, defined before aggregate reporting. \\
Baselines & local-only, one-shot LLM, periodic LLM, rule-based LLM, and learned-CVI policies. \\
learned-CVI threshold & \(\tau_{\mathrm{lite}}=0.997\), selected on validation logs before final evaluation. \\
Utility labels & \(K=5\) logged steps with token .10, latency .05, safety .15, and planner .05 weights. \\
Resource accounting & issued reasoner requests for Calls; PMR token-accounting field for Tokens. \\
\bottomrule
\end{tabular}
\end{table}

\subsection{Scenario Labels and Log Identifiers}
\label{app:world-labels}

The main text uses descriptive scenario labels. Table~\ref{tab:world-label-map}
reports the corresponding internal log identifiers used in the logged evidence
artifacts.

\begin{table}[!t]
\centering
\small
\caption{Publication scenario labels and internal log identifiers.}
\label{tab:world-label-map}
\begin{tabular}{lll}
\toprule
Publication label & Regime & Internal log ID \\
\midrule
Empty & Nominal & \texttt{W0\_empty} \\
Static Obstacles & Nominal & \texttt{W1\_static} \\
Mild Random Obstacles & Nominal & \texttt{W2\_random\_default} \\
Recoverable Dead End & Nominal & \texttt{W4\_deadend} \\
Multi-Obstacle & Hard/ambiguous & \texttt{W3\_multi} \\
Hard Re-entry A & Hard/ambiguous & \texttt{WH46\_test\_hard\_reentry} \\
Hard Re-entry B & Hard/ambiguous & \texttt{WH49\_test\_hard\_reentry} \\
Dead-end Bypass & Hard/ambiguous & \texttt{WM17\_left\_deadend\_right\_bypass} \\
\bottomrule
\end{tabular}
\end{table}

\subsection{Lower PPO Policy Details}
\label{app:ppo-lower-policy}

The lower PPO policy serves as the fixed local execution policy shared by all
invocation policies. It is trained before the agentic-recovery experiments in a
Gymnasium-based local-control surrogate environment.
Table~\ref{tab:ppo-lower-policy-config} summarizes the
implementation settings. The reward used for this lower policy encourages
local progress and terminal completion and penalizes collision, altitude
violation, repeated hold/inspection behavior, local-repair abuse, and
low-progress macros. Every selected macro skill passes through the same PMR
safety shield, verifier, and executor path used by agent-returned recovery
actions.

\begin{table}[!t]
\centering
\small
\caption{Fixed lower PPO policy configuration. This policy supplies the local
mission policy shared by all PMR invocation policies.}
\label{tab:ppo-lower-policy-config}
\begin{tabular}{lp{0.68\linewidth}}
\toprule
Item & Value \\
\midrule
PPO implementation & Stable-Baselines3 PPO, \texttt{MlpPolicy} \\
Network & two hidden layers, [64, 64], with SB3 default activation \\
Observation & 10 normalized local-flight features: front/left/right clearance,
goal distance, goal bearing, path-blocked flag, no-progress state,
repair-failure state, absolute altitude, and vertical error to goal altitude \\
Action space & 8 flying-phase tools $\times$ 4 repeats =
\(\mathrm{Discrete}(32)\); action decoded as tool index and repeat count \\
Repeat semantics & repeat \(r\in\{1,2,3,4\}\) means repeated sub-step/tool
calls, not seconds \\
Flying-phase tools & eight macro tools covering goal following, diagonal
altitude adjustment, local repair, scan/inspection, vertical motion, and safe hold \\
Runtime-owned actions & takeoff, landing, RTL, terminal homing, final approach,
completion, emergency handling, platform setpoints, and flight-stack primitives \\
Runtime ownership & PMR owns takeoff, completion, emergency handling, safety shield,
verifier, and platform-adapter execution; the reasoner selects only recovery
skills \\
Training worlds & \texttt{W0\_empty}, \texttt{W1\_static},
\texttt{W2\_random\_default}, \texttt{W3\_multi} \\
Key hyperparameters & learning rate \(3\times10^{-4}\), \(n_{\mathrm{steps}}=512\),
batch size 64, \(n_{\mathrm{epochs}}=10\), \(\gamma=0.99\),
\(\lambda_{\mathrm{GAE}}=0.95\), clip range 0.2, entropy coefficient 0.01,
4 vector environments, seed 42, 300k SB3 timesteps, max episode length 160 \\
Deployment & saved checkpoint loaded for deterministic inference and used
unchanged during local-only, one-shot, periodic, rule-based, or learned-CVI
evaluation \\
\bottomrule
\end{tabular}
\end{table}

\clearpage
Table~\ref{tab:ppo-lower-policy-sanity} reports a local-policy sanity check over
30 episodes per training world. This check verifies that the fixed local policy is
competent enough to serve as the default PMR execution path. The main method
comparison and learned-CVI threshold selection use the separate fixed protocol
described above.

\begin{table}[!t]
\centering
\small
\caption{Lower PPO policy sanity evaluation. The main PMR comparison keeps this
local policy fixed across all invocation policies.}
\label{tab:ppo-lower-policy-sanity}
\begin{tabular}{lrrr}
\toprule
Scenario & Runs & Rate & Mean return \\
\midrule
\texttt{W0\_empty} & 30 & 1.000 & 62.47 \\
\texttt{W1\_static} & 30 & 1.000 & 61.76 \\
\texttt{W2\_random\_default} & 30 & 0.967 & 60.34 \\
\texttt{W3\_multi} & 30 & 1.000 & 63.18 \\
\bottomrule
\end{tabular}
\end{table}

\subsection{Rule-Based LLM Baseline Details}
\label{app:rule-based}

Rule-based LLM is evaluated at each PMR runtime policy update instead of by the
fixed periodic due timer used by periodic LLM. The nominal periodic LLM due
interval is 5s, but actual calls are lower because requests are subject to the
shared budget, pending-recovery, terminal-suppression, cooldown, parser,
safety-shield, verifier, and executor path. Rule-based LLM proposes LLM re-entry
from expert-specified runtime events and then uses the same verifier/executor
path as learned-CVI. Table~\ref{tab:rule-based-details} reports the disclosed
rule structure used for the strong expert baseline.

\begin{table}[!t]
\centering
\small
\caption{Rule-based LLM trigger structure. The rule baseline is disclosed as the
strongest expert-specified comparison. Periodic LLM uses a fixed periodic due
timer and invokes the same reasoner when the shared runtime guards allow it.}
\label{tab:rule-based-details}
\begin{tabular}{p{0.43\linewidth}p{0.49\linewidth}}
\toprule
Runtime condition & Rule response \\
\midrule
Verifier reports completion & complete the mission. \\
MAVSDK unavailable in last result & hold safely and avoid unavailable commands. \\
Collision, safety abort, or battery below 20\% & trigger runtime-owned emergency response. \\
Unsafe proximity or front distance below 0.35m & hold safely near obstacles. \\
Within terminal radius with clear path and front distance at least 1.2m & continue local execution and suppress near-goal calls. \\
Ambiguity-loop mode with budget available & issue a synchronous LLM recovery call. \\
Front distance below repair trigger and repair failures below 3 & try local repair first. \\
No progress above 10s or repair failures at least 5, with budget & issue a synchronous LLM call for persistent local failure. \\
No progress above 5s or repair failures at least 3, with budget & issue an asynchronous LLM call as an early recovery request. \\
No progress above 15s but call unavailable & hold safely with local fallback. \\
Otherwise & continue local execution. \\
\bottomrule
\end{tabular}
\end{table}

The repair trigger depends on the clearance profile: 1.2m for safe, 0.9m for
balanced, and 0.8m for aggressive settings. In the reported comparison, the
important fairness point is that rule-based LLM and learned-CVI share the
reasoner, prompt contract, safety shield, verifier, executor, budget guards, and
cooldown machinery; the compared component is the trigger that proposes
re-entry.

\clearpage
\subsection{Latency and Backend Diagnostic}
\label{app:latency-backend}

Table~\ref{tab:latency-backend-diagnostic} reports the logged episode-level
latency scale for policies that issue remote-agent calls. The main resource
claim is therefore not that learned-CVI makes individual remote calls faster;
rather, it reaches the reported success/resource trade-off by admitting fewer
calls under the same remote-backend and verifier path.

\begin{table}[!t]
\centering
\small
\caption{Latency/backend diagnostic for episodes with at least one remote-agent
call in the final 400-run logs.}
\label{tab:latency-backend-diagnostic}
\begin{tabular*}{\linewidth}{@{\extracolsep{\fill}}lrrrrr}
\toprule
Method & Query eps. & Calls & Mean lat. (s) & Median lat. (s) & Backend fail \\
\midrule
one-shot LLM & 80 & 80 & 4.56 & 3.95 & 3 \\
periodic LLM & 80 & 240 & 5.05 & 4.25 & 15 \\
rule-based LLM & 80 & 96 & 4.89 & 4.41 & 4 \\
learned-CVI & 45 & 46 & 4.69 & 4.27 & 1 \\
\bottomrule
\end{tabular*}
\end{table}

\subsection{Sparse Invocation Ablation Notes}
\label{app:sparse-invocation-diagnostics}

Table~\ref{tab:sparse-invocation-diagnostics} in Section~\ref{sec:result}
reports the closed-loop sparse-invocation ablation counts. The diagnostic
separates the learned score, the hard-stuck guard, and a matched-budget
rule-based comparison under the same runtime contract. The primary task metric
remains the Clean Success@1m comparison in Table~\ref{tab:fixed-protocol-eval};
backend timeouts are counted by the same Clean Success@1m criterion.

\subsection{CVI Data Size and Model-Class Diagnostic}
\label{app:cvi-model-diagnostic}

The learned-CVI model is intentionally compact. It uses the fixed 18D runtime
vector in Table~\ref{tab:cvi-features}, excluding scenario identifiers, map
templates, raw range or perception arrays, prompt geometry, token text, and raw
prompts. The short-horizon label set contains 651 labeled training samples after
filtering, motivating a sigmoid-linear gate with transparent runtime behavior.

We also ran a model-class diagnostic because onboard UAV deployment favors
admission gates no heavier than small MLP-scale models.
The diagnostic compares learned-CVI with a small MLP and a gradient-boosted-stump
ensemble using the same 18D features and labels.
Table~\ref{tab:cvi-model-class-diagnostic} reports threshold-free ranking
metrics: ROC-AUC and PR-AUC. The nonlinear models fit validation labels
strongly, while learned-CVI keeps stronger held-out ROC-AUC under the same
feature contract. We therefore use the linear learned-CVI gate because it
combines favorable ranking behavior with transparent weights and low runtime
overhead.

\begin{table}[!t]
\centering
\small
\caption{Threshold-free learned-CVI model-class diagnostic using the same 18D
features and short-horizon labels.}
\label{tab:cvi-model-class-diagnostic}
\begin{tabular*}{\linewidth}{@{\extracolsep{\fill}}llcc}
\toprule
Split & Model & ROC-AUC & PR-AUC \\
\midrule
Validation & learned-CVI & 0.9996 & 0.9996 \\
Validation & small MLP [32,16] & 1.0000 & 1.0000 \\
Validation & boosted decision stumps & 1.0000 & 1.0000 \\
\midrule
Held-out diagnostic & learned-CVI & 0.9635 & 0.9986 \\
Held-out diagnostic & small MLP [32,16] & 0.6968 & 0.9861 \\
Held-out diagnostic & boosted decision stumps & 0.2976 & 0.9610 \\
\bottomrule
\end{tabular*}
\end{table}

\clearpage
\subsection{Data-Size and Utility-Weight Diagnostics}
\label{app:utility-sensitivity}

\begin{center}
\small
\refstepcounter{table}\label{tab:cvi-data-size-diagnostic}
\textbf{Table~\thetable:} Log-spaced learned-CVI data-size diagnostic.
\vspace{4pt}

\begin{tabular*}{0.72\linewidth}{@{\extracolsep{\fill}}rrr}
\toprule
Training rollouts & Labeled rows & Cosine sim. \\
\midrule
8  & 111 & 0.8037 \\
16 & 282 & 0.9888 \\
32 & 651 & 1.0000 \\
48 & 651 & 1.0000 \\
\bottomrule
\end{tabular*}
\end{center}

\vspace{-2pt}
\begin{center}
\small
\refstepcounter{table}\label{tab:cvi-utility-sensitivity}
\textbf{Table~\thetable:} Utility-weight sensitivity diagnostic.
\vspace{4pt}

\begin{tabular*}{\linewidth}{@{\extracolsep{\fill}}lrrr}
\toprule
Regime & Grid & Label agreement & Utility-positive queries \\
\midrule
Nominal & 729 & 0.9985 [0.9956, 1.0000] & 1.000 / 1.000 / 1.000 \\
Hard/ambiguous & 729 & 0.9682 [0.9371, 1.0000] & 1.000 / 1.000 / 1.000 \\
\bottomrule
\end{tabular*}
\end{center}

We inspected a log-spaced data-size diagnostic over training rollout counts.
The learned-CVI weight direction stabilizes rapidly as the training pool grows,
reaching 0.989 cosine similarity to the full-pool gate by 16 rollouts and
matching the full-pool direction by 32--48 rollouts. This serves as a
data-efficiency diagnostic for the compact gate.

We run a sensitivity diagnostic over 729 short-horizon recovery-utility
configurations varying the horizon, utility threshold, and token, latency,
safety-rejection, and planner-failure weights. The diagnostic recomputes labels
and checks the states where learned-CVI invoked the reasoner in existing logs.
Across the grid, label agreement with the baseline utility remains high, and
all executed query states remain utility-positive under every tested
configuration. This supports robustness of the reported triggered states to
reasonable utility-weight changes. The aggregate diagnostic is reported in
Table~\ref{tab:cvi-utility-sensitivity}.

Table~\ref{tab:per-world-learned-cvi} in Section~\ref{sec:result} reports the
per-scenario decomposition of learned-CVI behavior, and
Table~\ref{tab:cvi-threshold-diagnostic} reports the learned-CVI threshold
diagnostic. These auxiliary diagnostics characterize the recorded learned-CVI
score behavior around the deployed operating point.

Failure-pattern diagnostics are consistent with the main comparison:
learned-CVI has zero nominal non-successes and two hard/ambiguous
non-successes, corresponding to one threshold-edge under-query and one backend
timeout after admission. Local-only hard/ambiguous non-successes concentrate in
dead-end and branch-trap settings, while rule-based LLM's single hard/ambiguous
non-success is associated with backend exposure under the same Clean
Success@1m criterion.

\subsection{Real-Platform Demonstration}
\label{app:real-platform}

For the real-platform demonstration, PMR/OpenClaw runs on a laptop while a
Crazyflie nano-quadcopter executes the mission through the constrained offboard
recovery interface. The demonstration uses a blocked Dead-end Bypass-style
layout: local execution first enters a blocked branch, learned-CVI admits a
recovery query when blocked or no-progress evidence accumulates, and the
accepted recovery skill is routed back through the PMR executor boundary before
changing the Crazyflie command stream. The main text reports ten repeated
trials for local-only and learned-CVI. The accompanying waypoint exports and
visualization files document the physical path geometry and blocked-navigation
setup. This artifact provides real-platform evidence for the PMR invocation and
execution boundary, while the aggregate performance claims are based on the
400-run simulation benchmark.

\end{document}